\documentclass[letterpaper]{article} 
\usepackage{aaai25}  
\usepackage{times}  
\usepackage{helvet}  
\usepackage{courier}  
\usepackage[hyphens]{url}  
\usepackage{graphicx} 
\usepackage{natbib}  
\usepackage{caption} 
\usepackage{algorithm}
\usepackage{algorithmic}

\usepackage{amsfonts}
\usepackage{bm}
\usepackage{amsmath}
\usepackage{booktabs}
\usepackage{blindtext}
\usepackage{xcolor}
\usepackage{nameref}

\usepackage{newfloat}
\usepackage{listings}
\DeclareCaptionStyle{ruled}{labelfont=normalfont,labelsep=colon,strut=off} 
\floatstyle{ruled}
\newfloat{listing}{tb}{lst}{}
\floatname{listing}{Listing}
\usepackage[breaklinks,colorlinks]{hyperref}

\def\name{GaussianPainter }
\def\namenospace{GaussianPainter}
\title{\namenospace: Painting Point Cloud into 3D Gaussians with Normal Guidance} 
\author{
    Jingqiu Zhou\textsuperscript{\rm 2}\thanks{Equal contribution.},
    Lue Fan\textsuperscript{\rm 2,3,4}\footnotemark[1],
    Xuesong Chen\textsuperscript{\rm 2},
    Linjiang Huang\textsuperscript{\rm 1}\thanks{Corresponding author.},
    Si Liu\textsuperscript{\rm 1},
    Hongsheng Li\textsuperscript{\rm 2,4}
}
\affiliations{
    \textsuperscript{\rm 1}Beihang University\\
    \textsuperscript{\rm 2}Multimedia Laboratory, The Chinese University of Hong Kong\\
    \textsuperscript{\rm 3}Chinese Academy of Sciences\\
    \textsuperscript{\rm 4}Centre for Perceptual and Interactive Intelligence\\
}

\usepackage{bibentry}

\begin{document}

\maketitle

\begin{figure*}[!t]
  \centering
  \includegraphics[width=\linewidth]{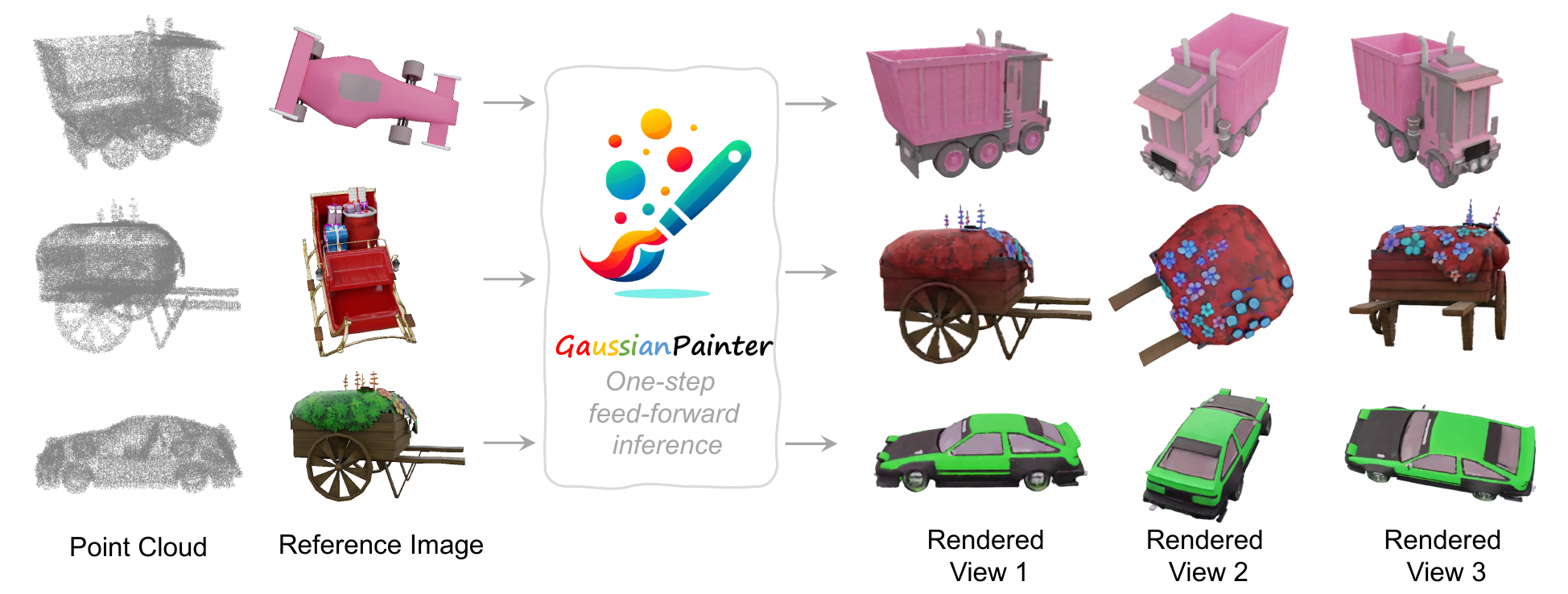}
    \vspace{-5mm}
  \caption{Given any reference image, the \name paints point clouds into 3D Gaussians in a feed-forward network.
  }
  \vspace{-5mm}
  \label{fig:teaser}

\end{figure*}

\begin{abstract}
In this paper, we present \namenospace, the first method to paint a point cloud into 3D Gaussians given a reference image.
\name introduces an innovative feed-forward approach to overcome the limitations of time-consuming test-time optimization in 3D Gaussian splatting. Our method addresses a critical challenge in the field: the \emph{non-uniqueness} problem inherent in the large parameter space of 3D Gaussian splatting. This space, encompassing rotation, anisotropic scales, and spherical harmonic coefficients, introduces the challenge of rendering similar images from substantially different Gaussian fields. As a result, feed-forward networks face instability when attempting to directly predict high-quality Gaussian fields, struggling to converge on consistent parameters for a given output.
To address this issue, we propose to estimate a surface normal for each point to determine its Gaussian rotation.
This strategy enables the network to effectively predict the remaining Gaussian parameters in the constrained space. We further enhance our approach with an appearance injection module, incorporating reference image appearance into Gaussian fields via a multiscale triplane representation.
Our method successfully balances efficiency and fidelity in 3D Gaussian generation, achieving high-quality, diverse, and robust 3D content creation from point clouds in a single forward pass. Our code is available at \url{https://github.com/zhou745/GaussianPainter}.
\end{abstract}

\section{Introduction}\label{sec_intro}
3D object generation has attracted increasing attention due to the potential to reduce human labor and professional software in the game, VR, and animation industries.

Over the past few years, implicit representations such as NeRF~\cite{nerf,xu2022point,kondo2021vaxnerf} and Signed Distance Function (SDF)~\cite{sdf,park2019deepsdf} have played major roles in this field.
Recently, the emerging 3D Gaussian Splatting (3DGS)~\cite{gsp} has become a hot trend because of its impressive rendering quality and efficiency.
Different from implicit representations, 3D Gaussians can be viewed as a special type of point cloud with each point being decorated by rotation, scales, and colors.
Such affinity between the Gaussians and point clouds raises a natural question: \emph{Given reference information (e.g., reference images), can we transform a point cloud into a 3D Gaussian field in an efficient feed-forward manner?} 
Here the point clouds can be obtained from the existing 3D assets or point cloud generative models, such as Point-E~\cite{nichol2022point}, PointFlow~\cite{yang2019pointflow}, and PointGrow~\cite{sun2020pointgrow}.
In this way, we try to take advantage of numerous existing 3D assets and the mature point cloud generators, and then efficiently \emph{paint} point clouds into 3D Gaussians, as indicated by the paper title.

Compared with our method, existing methods~\cite{chen2023text, tang2023dreamgaussian, haque2023instruct} for 3D Gaussian generation are based on multi-step optimization or denoising, instead of a feed-forward paradigm.
These methods are inefficient due to their multiple forward and backward iterations.
Although the feed-forward Gaussian painting is more efficient, it is quite challenging for the following reason.
3D Gaussians are designed with a relatively large parameter space, unlike the previous point-based rendering methods~\cite{lassner2021pulsar,yifan2019differentiable} only utilizing spheres without rotations and anisotropic scales.
Although such a large parameter space leads to the \emph{non-uniqueness} of Gaussian fields.
In other words, similar images can be rendered from totally different Gaussian fields, making it ambiguous for feed-forward models to predict Gaussian parameters in a single forward pass.
This issue is also observed by recent work AGG~\cite{xu2024agg}, which turns to \emph{isotropic} Gaussians (i.e., ignore rotations) to constrain the parameter space for more stable training and easier prediction. 
However, this solution weakens the capacity of 3D Gaussian and rendering quality.
\par
In this paper, we propose \namenospace, which takes a point cloud and a reference image as input, generating 3D-Gaussian parameters for each point in a single forward pass.
\name contains two major components including (1) \emph{Normal-guided Gaussian Painting} (Sec.~\ref{sec:gs_prediction}) and (2) \emph{Triplane-based Multiscale Appearance Injection} (Sec.~\ref{sec:injection}).
The first normal-guided Gaussian painting part addresses the issue of non-unique Gaussian fields by introducing \emph{normal guidance}, where surface normals are treated as guidance to constrain the parameter space of Gaussians.
More specifically, we first propose \emph{Isotropic Normal Rendering} module to estimate a surface normal for each input point.
The estimated normal is leveraged to define the rotations of the Gaussian centered at the point.
After obtaining the rotations, the space of Gaussian parameters is compressed and constrained.
It is much easier for the neural network to predict the remaining parameters (i.e. scales, opacity, and colors).
Moreover, the \emph{Triplane-based Multiscale Texture Injection} part then injects the appearance from the reference image to Gaussians, for controllable Gaussian painting.
In this module, the triplane representation narrows the modality gap between 2D reference images and 3D points, and the multiscale feature enables high-fidelity appearance injection. 
\par
To summarize, our contributions are of three folds:
\begin{enumerate}
    \item We treat the problem of 3D Gaussian generation as painting given point clouds into 3D Gaussians with a feed-forward neural network.
    This problem setup offers 3D object generation from a new perspective and leads to an efficient tool for creating diverse 3D content. 
    \item We analyze and effectively address the issue of non-unique Gaussian fields by introducing normal guidance on the training process, making the efficient feed-forward manner feasible.
    \item Our method achieves state-of-the-art performance on the novel view synthesis task. In addition, we propose a novel cross-object appearance transfer task and our method demonstrates generalized transferring ability.
    
\end{enumerate}
\begin{figure*}[tb]
  \centering
  \includegraphics[width=0.9\linewidth]{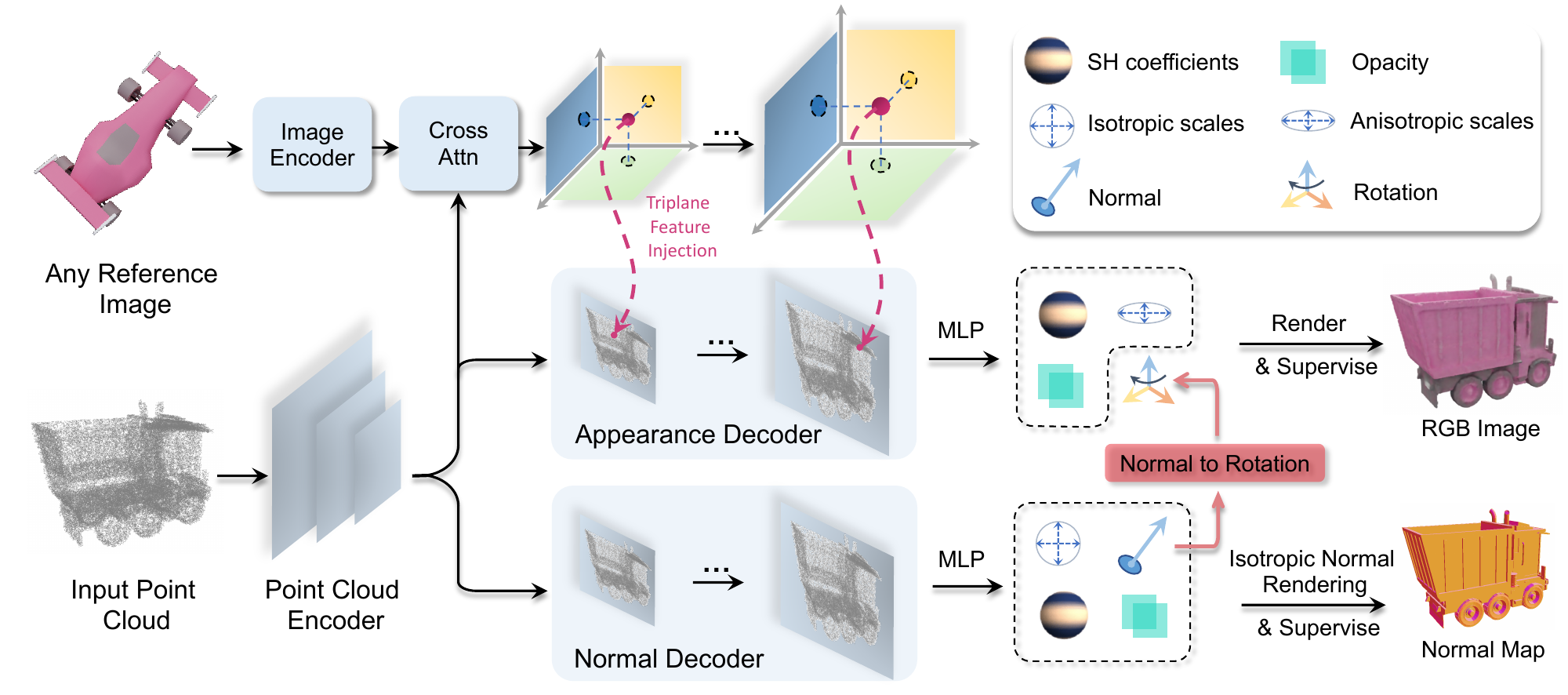}
  \caption{
  The overview of \namenospace.
  Its first major component is \emph{Normal-guided Gaussian Painting}, consisting of Point Cloud Encoder, Appearance/Normal Decoder, and the following MLPs. This component directly operates on point clouds and predicts Gaussian parameters and normal for each point, presented in Sec.~\ref{sec:gs_prediction}.
  The second major component is \emph{Triplane-based Multiscale Appearance Injection}, which injects appearance information from reference image into the Appearance Decoder and guides Gaussian painting, presented in Sec.~\ref{sec:injection}.
  }
  \label{fig:pipe}
  \vspace{-3mm}
\end{figure*}

\section{Related Work}
\subsection{3D Gaussian Splatting}
3D Gaussians~\cite{gsp} is an explicit 3D representation that encodes rendering information in 3D anisotropic Gaussians. For each 3D Gaussian, it has the following properties: the mean $\mu\in \mathbb{R}^3$, the covariance $\Sigma \in \mathbb{R}^{3\times 3}$, a scalar $\sigma$ for opacity, and the view-dependent spherical harmonic coefficients $c\in\mathbb{R}^n$. The 3D Gaussian splitting has become a hot trend and is employed by many methods~\cite{qian20233dgs, luiten2023dynamic, yan2024street, guedon2023sugar,gao2023relightable,zhou2023drivinggaussian,yugay2023gaussian} for various 3D tasks.
Among them, a line of work~\cite{chen2023text, tang2023dreamgaussian, li2023gaussiandiffusion, liang2023luciddreamer, liu2023humangaussian, chung2023luciddreamer} focuses on the Gaussian generation from text instructions.
Although these methods show promising results in generating objects, they usually take minutes to generate a single object due to the multi-step denoising process.
In contrast, some methods~\cite{tri2gsp,xu2024agg} adopt feed-forward inference to create 3D objects from a single image, which is much faster than diffusion-based methods.
However, the feed-forward manner faces a critical challenge of non-uniqueness when predicting 3D Gaussians, which will be presented in Sec.~\ref{sec:pilot}.
In this paper, we are dedicated to boosting the performance of feed-forward methods by tackling this challenge.


\subsection{3D Object Generation}
Generative AI leverages machine learning models to create new content such as text, images~\cite{he2024freeedit,huang2025fouriscale}, videos~\cite{zhang2023controlvideo}, and 3D objects, has seen rapid advancements in recent years. Among them, 3D object generation has always posed significant challenges. Earlier approaches~\cite{li2021mine, shih20203d, xu2022sinnerf} focused on limited view synthesis and achieved impressive 3D consistency.
The recent advancements in 2D diffusion models lead to the development of a new diffusion-based paradigm~\cite{poole2022dreamfusion,wang2023score,wang2024prolificdreamer} for 3D object generation. In this pipeline, the 2D diffusion model serves as a prior, and the 3D content is generated through optimization guided by this prior.
This paradigm leads to many previous arts~\cite{lin2023magic3d, liu2023zero1to3, liu2024one, melas2023realfusion}.
Similar to the diffusion-based generation method in 3D Gaussian, these methods also face the inefficient multi-step denoising process, which again encourages us to explore the feed-forward methods.

\section{Pilot Study: Understanding the Non-uniqueness of 3D Gaussians}
\label{sec:pilot}

As mentioned in Sec.~\ref{sec_intro}, similar images can be rendered from totally different Gaussian fields, which makes it difficult for a generative model to predict high-quality 3D Gaussians. 
In this section, we conduct a pilot study to demonstrate this issue.
Particularly, we first load a well-trained Gaussian field and fix the locations of each 3D Gaussian.
We then randomly re-initialize and tune other Gaussian parameters, including rotation, scale, opacity, and spherical harmonics (SH) coefficients.
This process is repeated multiple times, obtaining multiple Gaussian fields with the same locations but different other parameters.
For each type of parameter, we define an Instability Score (IS) to measure its instability.
\par
Assuming that we repeatedly train $M$ Gaussian fields with $N$ Gaussian locations in each field, the IS of a parameter with $C$ channels is calculated by following steps.
(1) For each location and each channel, we calculate the standard variance of the $M$ scalar values (one value per field).
Thus, in total, we have $N \times C$ standard variances. We define their mean as \emph{local instability score}.
(2) Since different types of parameters have different scales, we calculate a global parameter scale for each type to normalize the instability score. For a certain parameter, its global parameter scale is defined as the standard variance of all the $M \times N \times C$ scalar values.
(3) The final instability score is the ratio between the local instability and the global parameter scale.
By definition, a large instability score indicates a large degree of non-uniqueness.
\par
\begin{figure}[ht]
  \centering
  \includegraphics[height=4.7cm]{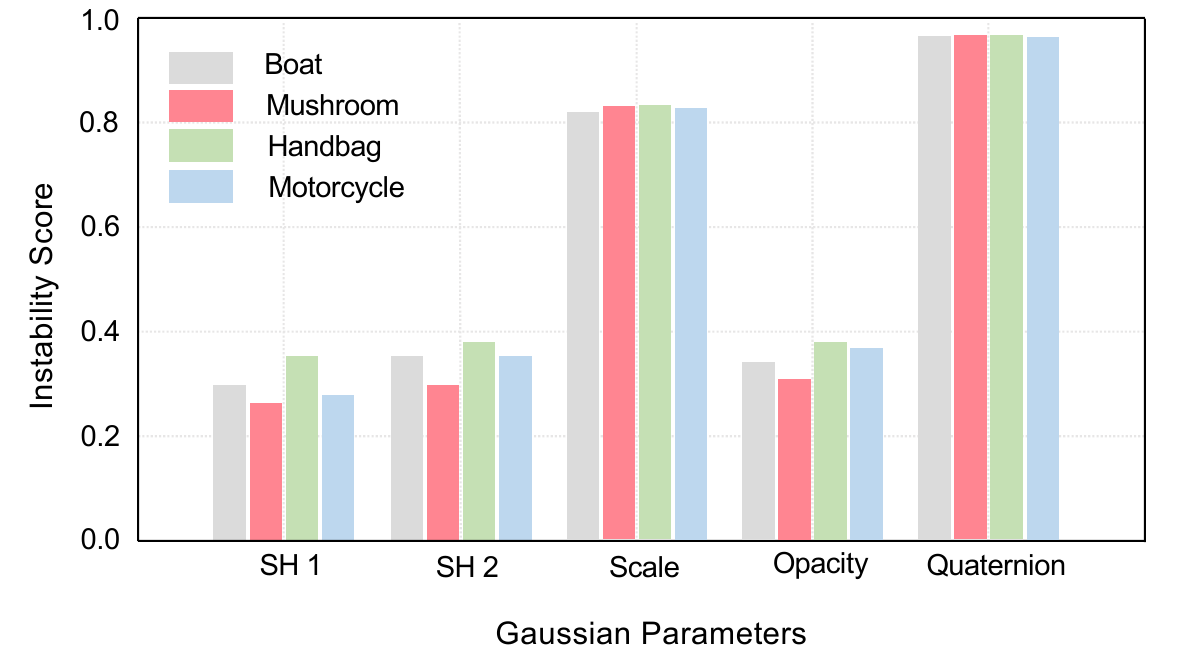}
    \vspace{-10pt}
  \caption{Demonstration of different Gaussian fields and Instability Score for different Gaussian parameters.
  SH 1 and SH 2 stand for the RGB component and the remaining component of harmonic coefficients, respectively.}
  \label{fig:ngp}
  \vspace{-5pt}
\end{figure}

As shown by the statistics in Fig.~\ref{fig:ngp}, SH and opacity are relatively stable between different re-initializations.
In contrast, scales and rotations are quite unstable, which are difficult to be predicted by generative models.
Since scales are entangled with rotations, we consider the prediction of stable rotations as a key challenge to tackle, leading to our motivation to propose Normal Guidance to make the rotation prediction more stable and ``unique''.



\section{Method}

The proposed \name takes a point cloud and a reference image as input, generating 3D-Gaussian parameters for each point in a single forward pass, including SH coefficients, opacity, scales, and rotations.
The generated 3D Gaussians are expected to render high-quality images with a similar appearance provided by the reference image.
\name contains two major components: (1) \emph{Normal-guided Gaussian Painting} (Sec.~\ref{sec:gs_prediction}) and (2) \emph{Triplane-based Multiscale Appearance Injection} (Sec.~\ref{sec:injection}).
The former introduces normal guidance to address the issue of non-unique Gaussians, and the latter injects appearance information from the reference image into 3D Gaussians.
Fig.~\ref{fig:pipe} illustrates the overall architecture of \namenospace.

\subsection{Normal-guided Gaussian Painting}
\label{sec:gs_prediction}
As discussed in the pilot study, the difficulty of Gaussian parameter prediction lies in the non-uniqueness and large parameter space of 3D Gaussians.
In this section, we propose a normal guidance technique to constrain the parameter space for achieving more stable and better prediction.

\subsubsection{Point Cloud Encoding.}
Given a point cloud (either from 3D scanning or generated by neural networks), we start by converting it to an occupancy grid with dimensions of $200 \times 200 \times 200$.
We then adopt a 3D UNet for encoding the input occupancy.
The 3D UNet is adapted from the conventional 2D UNet~\cite{unet}, where all 2D convolution layers are replaced with 3D sparse convolutions.
In addition, all Batch Normalization layers are replaced with Layer Normalization to alleviate the unstable batch statistics caused by the sparsity of point clouds. 
The decoders of the 3D UNet upsamples the occupancy grid back to its original resolution.
Note that we have two UNet decoders sharing the same structure as shown by Fig.~\ref{fig:pipe}. The normal decoder predicts normals and other parameters within an isotropic Gaussian setting, while the appearance decoder predicts all parameters within an anisotropic setting for RGB rendering. We will present the details later.

\subsubsection{Isotropic Normal Rendering.}
Based on the extracted occupancy features, we employ a multi-layer perceptron (MLP) to learn a directional vector for each occupied position.
This vector is expected to point to the normal direction of the closest surface.
In practice, some positions may deviate from the actual surface, due to the inherent noise in raw point clouds and the introduced extra occupied positions in the Point Cloud Encoding.
As a result, defining precise normals for these off-surface positions is challenging. To address this issue, we refrain from directly supervising normal estimation in 3D space.
Instead, we propose \emph{Isotropic Normal Rendering} to render surface normals to a normal map and employ supervision on the 2D normal map.
Specifically, an MLP takes the encoded occupancy feature as input and predicts Gaussian parameters for each occupied position, including a normal, isotropic scale, and opacity.
Then we render the normals into a 2D normal map following
\begin{equation}
    \bm{n} = \sum_{i = 0}^M \bm{n}_i\alpha_i\prod_{j=1}^{i-1}(1 - \alpha_j).
    \label{eq:normal_render}
\end{equation}
Eq.~\eqref{eq:normal_render} substitutes colors in the original 3DGS rendering with predicted normals.
$\alpha$ is calculated from the predicted opacity and Gaussian density.
Since the issue of non-uniqueness still exists during the normal prediction, the Gaussians in this module are defined as \emph{isotropic}, to facilitate the prediction.
Finally, the rendered normal map is supervised by L1 loss and SSIM loss.
The properties of the proposed Isotropic Normal Rendering are worth further discussion:

\begin{itemize}
    \item Leveraging normal rendering allows for the approximation of normals for all 3D positions, including those positions deviating from actual surfaces.
    \item Although these normals may not precisely match the true surface normals, they are sufficient to define rotations and constrain the space of feasible Gaussian parameters, fulfilling our requirements.
    \item The Gaussians here only serve the purpose of normal rendering.  With the learned normals, a \emph{new} set of Gaussians will be generated, which are anisotropic rather than isotropic, which will be discussed later.
\end{itemize}

\subsubsection{Gaussian Prediction with Normal Guidance.}
\begin{figure}[tb]
  \centering
  \includegraphics[width=0.9\linewidth]{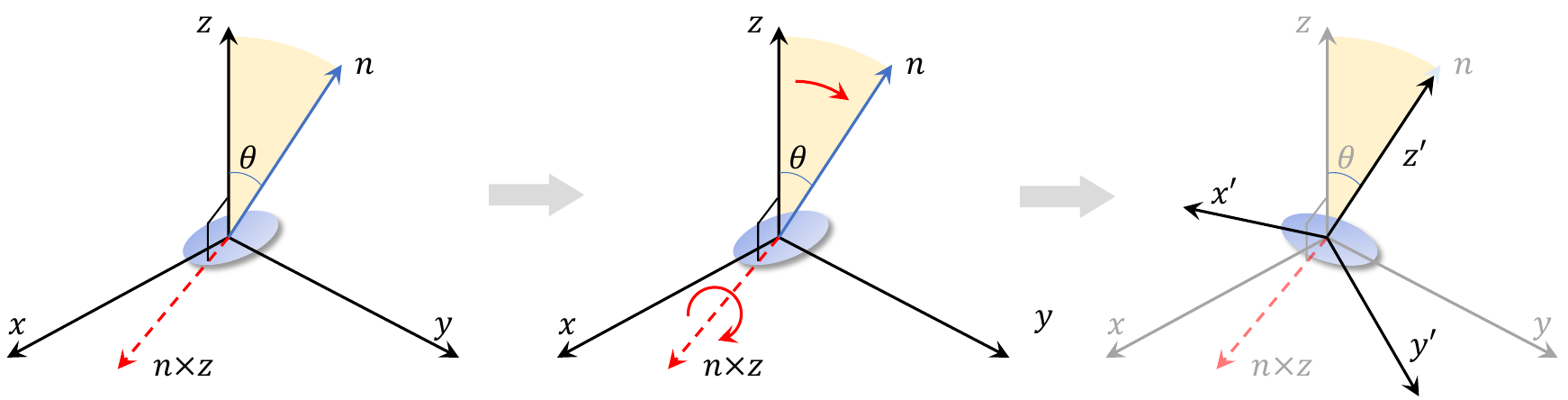}
  \caption{The illustration of converting the predicted normal to a rotation. The $xyz$-axes are rotated to $x^\prime y^\prime z^\prime$-axes, which is considered as the rotation parameter of a Gaussian.}
  \label{fig:n2q}
\vspace{-10pt}
\end{figure}

Each occupied position is assigned a predicted normal $\bm{n}$, which is assumed to be a unit vector.
We then demonstrate how to specify the Gaussian rotation $\bm{R} \in \mathbb{R}^{3\times 3}$ according to $\bm{n}$.
Considering the three axes of the local coordinate system of a position, 
these axes are rotated from the canonical pose to pose $\bm{R}$, around a designated rotation axis $\bm{r}$.
The rotation axis $\bm{r}$ is designated as
\begin{equation}
    \bm{r} = \bm{n} \times \bm{z},
\end{equation}
where $\bm{z}$ is the unit vector along $z$-axis of the world coordinate (vertical direction).
The rotation angle $\theta$ is the angle between $\bm{z}$ and $\bm{n}$.
With rotation axis $\bm{r}$ and rotation angle $\theta$, rotation $\bm{R}$ is calculated using Rodrigues' rotation formula, the details of which are omitted for brevity.
Fig. \ref{fig:n2q} illustrates the rotation procedure.

\par
After obtaining the rotation $\bm{R}$, we leverage the appearance decoder and an MLP to predict the remaining Gaussian parameters based on the occupancy features.
There are two notable differences from the Isotropic Normal Rendering module.
1) Here the MLP predicts a scale for each axis independently, so we have anisotropic Gaussians rather than isotropic Gaussians adopted in Normal Rendering.
2) The MLP predicts new opacity without reusing the opacity in the Normal Rendering module.
With the newly predicted Gaussians, we render them into images and employ L1 loss and SSIM loss to optimize all trainable parameters in our model.

\subsection{Triplane-based Multiscale Appearance Injection}
\label{sec:injection}
To control the Gaussian painting, we propose Appearance Injection to inject the appearance information from reference images into the Gaussian painting process.
This injection module features a multi-scale triplane design.
The triplane representation bridges the dimensional gap between 3D Gaussians and 2D reference images, making the appearance injection aware of 3D structures.
Moreover, the multi-scale design facilitates preserving the fidelity of appearance, which will be experimentally verified later.

\subsubsection{Image Encoding with Multi-scale Triplane.}
Given a reference image, we first utilize DINOv2~\cite{oquab2023dinov2} to encode it into visual feature maps $\mathcal{F}$.
Then we enhance $\mathcal{F}$ with 3D information by a simple cross attention with 3D occupancy features, formulated as
\begin{equation}
    \mathcal{\Tilde{F}} = \texttt{CrossAttn}(\mathcal{F}, [\mathcal{F}, \mathcal{O}], [\mathcal{F}, \mathcal{O}]),
    \label{eq:attn}
\end{equation}
where $\mathcal{O}$ is the flattened occupancy features from the last layer of the point cloud encoder and $[\;]$ stands for concatenation in token dimension.
In Eq.~(\ref{eq:attn}), $\mathcal{F}$ serves as attention queries and $ [\mathcal{F}, \mathcal{O}]$ serves as keys and values.
Here, we omit the notation for the flatten operation on $\mathcal{F}$ for clarity.
We then denote the $i$-th plane of a triplane structure in the $s$-th scale as $\mathcal{T}_s^i$, where the superscript $i$ stands for plane direction and $i \in \{\text{xy}, \text{yz}, \text{zx}\}$. $\mathcal{T}_s^i$ is constructed with the following rules
\begin{equation}
    \mathcal{T}_s^i = 
    \begin{cases}
        \texttt{UpLayer}_i(\mathcal{\Tilde{F}}),\; \text{if } s = 0 \\ 
    \texttt{UpLayer}_i(\mathcal{T}_{s-1}^i),\; \text{if } s > 0
    \end{cases},
    \label{eq:triplane}
\end{equation}
where $\texttt{UpLayer}_i$ is an upsampling layer with stride 2 in 2D UNet~\cite{ldm}. The subscript $i$ denotes the upsample layers for three directions, each with unique parameters.

\subsubsection{Appearance Injection.}
With the multi-scale triplane features, we use each occupied 3D position to fetch the corresponding features in the three planes.
We choose $N_s$ layers in the appearance decoder of the UNet as \emph{injected layers}, where $N_s$ is the number of triplane scales.
We denote an occupied position in the injected layer with scale $s$ as $\bm{p}_s \in \mathbb{R}^3$.
Its fetched feature from the triplane $\mathcal{T}_s$ can be denoted as 
\begin{align}
    \bm{f}^{\bm{p}}_s = \texttt{CAT}(&\texttt{Interp}(\mathcal{T}_s^{\text{xy}}, \bm{p}^{\text{xy}}_s),\nonumber\\
    &\texttt{Interp}(\mathcal{T}_s^{\text{yz}}, \bm{p}^{\text{yz}}_s),\nonumber\\
    &\texttt{Interp}(\mathcal{T}_s^{\text{zx}}, \bm{p}^{\text{zx}}_s)),
\end{align}
where $\texttt{CAT}$ and $\texttt{Interp}$ stand for concatenation along channel dimension and bilinear interpolation, respectively.
To inject the triplane feature into the occupancy grid, $\bm{f}^{\bm{p}}_s$ is projected by a linear layer to have the same channel dimension with the occupancy feature of $\bm{p}$.
Then the projected $\bm{f}^{\bm{p}}_s$ is added to the occupancy feature. 
In this way, the multi-scale triplane features are injected into multiple layers of the point cloud decoder, guiding the rendering of Gaussians.

In conclusion, as shown in Fig.~\ref{fig:pipe}, the \emph{Normal Decoder} and \emph{Appearance Decoder} are pivotal to our method.
The normal decoder predicts normals for occupied positions under an isotropic Gaussian setting.
The Isotropic Normal Rendering module enables supervising normals on a 2D normal map.
The appearance decoder integrates appearance information from multi-scale triplanes and normal guidance from the normal decoder, ultimately predicting the anisotropic Gaussians as the final output of \namenospace.

\section{Experiments}

\subsection{Implementation Details}\label{imd}

\subsubsection{Point Cloud Encoder and Decoder.} 
Based on the basic structure of 2D UNet~\cite{ldm}, the point cloud encoder progressively downsamples point clouds (i.e., occupancy grids after preprocessing).
Each downsampling step halves the spatial size and doubles the channel size.
In particular, we downsample the input occupancy grid by a factor of 16, while increasing the channel dimension from 32 to 512.
Following the encoder, the normal decoder and appearance decoder gradually upsample the spatial size and decrease the channel dimension.
Specifically, the two decoders recover the spatial size by $16\times$ and reduce channel dimension from 512 to 32. 

\subsubsection{Image Encoder and Triplanes.}
DINOv2~\cite{oquab2023dinov2} with ViT-based~\cite{vit} backbone is adopted as the image encoder.
In our implementation, the DINOv2 encodes a $518 \times 518$ reference image into a $37\times37$ feature map. 
This feature map is further lifted into triplane features in four scales, including $37 \times 37$, $74 \times 74$, $144 \times 144$, and $288 \times 288$. Correspondingly, the channel dimensions of these triplanes are 512, 256,128, and 64.

\subsubsection{Training Scheme.}
We take a single object as an example to demonstrate our training scheme.
We first pre-render $K$ views of this object.
For each training iteration, we randomly sample a view as the reference image.
Our model then predicts a Gaussian field and normals based on the reference image and point cloud.
The predicted Gaussian field and normals are rendered into another random view for supervision.
It is necessary to emphasize that solely using rendering loss without the need to train a Gaussian field for each object greatly simplifies the workflow.

\subsection{Datasets, Evaluation, and Compared Methods}
\label{sec:data_eval}

\subsubsection{OmniObject3D}\cite{ommni3d} is a 3D dataset with over 6000 objects in 197 categories, which provides a sufficient number of blender-rendered views, surface normal maps, and point clouds (16384 points for each object).
A subset of OmniObject3D which consists of 2437 objects across 73 categories is used as our total dataset.
Within this subset, we sample a \emph{validation split} for the evaluation of novel view synthesis, which contains two objects for each category.

To train our model on this dataset, we adopt a two-stage training schedule spanning 12 epochs.
\emph{Stage 1:} During the first 8 epochs, we set the spherical harmonic degree to 0. This ensures fast convergence and stability.
\emph{Stage 2:} In the remaining 4 epochs, we gradually increase the spherical harmonic degree from 0 to 1. This adjustment enables the model to capture more complex appearances.
Throughout the entire training process, we use the Adam optimizer and maintain a constant learning rate of $1 \times 10^{-5}$.

\subsubsection{Objaverse}\cite{objaverse} is a large-scale 3D dataset containing over 80,000 renderable 3D models for Blender. From the LVIS split, we select 10,000 high-quality assets to construct our dataset. Unlike OmniObject3D, Objaverse does not include point clouds, rendered images, or surface normals. To generate realistic images, we render these objects using an HDRI environment and Blender's Cycles engine. Each object is rendered from 100 distinct poses, saving the corresponding surface normal maps. Point clouds are extracted using a surface sampling strategy~\cite{trimesh}, sampling 32,768 points per object due to the higher complexity of Objaverse models compared to OmniObject3D.

For training, we fine-tune the model pretrained on the OmniObject3D dataset for 24 epochs on the Objaverse subset. We use the Adam optimizer with a fixed learning rate of $1 \times 10^{-5}$ throughout the process.

\subsubsection{Evaluation.}
We conduct both quantitative and qualitative evaluations.
For better understanding, we assume each object has $K$ ground-truth views.
For the evaluation of the $i$-th view, we randomly choose another view from the remaining $K-1$ views as the reference to paint the point cloud into Gaussians.
Then the generated Gaussians are rendered into the $i$-the view to evaluate the quality of the $i$-th rendered view.
The process above is repeated $K$ times with a different reference view and rendering view each time, which avoids the bias caused by a fixed choice of reference view.
The overall evaluation result of each object is the average of results in $K$ views.
For qualitative evaluation of an object, we can specify any image as the reference, which is not limited to a certain rendered view of the object.
\subsubsection{Compared Methods.}
We compare our method with the two baseline methods AGG~\cite{xu2024agg} and TriplaneGaussian~\cite{tri2gsp}.
AGG is a two-stage method that first generates a reasonable point cloud according to the input image and then predicts Gaussian parameters based on the generated point cloud.
Given that AGG is not open-sourced, for a fair comparison, we substitute its generated point clouds with ground-truth point clouds and re-implement their pipeline, achieving similar results to their official results.
The most distinctive difference between our \name and AGG lies in that AGG simply predicts isotropic Gaussians, overlooking the challenge of non-uniqueness presented in Sec.~\ref{sec:pilot}.
\par
TriplaneGaussian is another related work to generate 3D Gaussians in a feed-forward manner.
The authors provide an official inference script and pretrained weights for the Objaverse dataset.
For a fair comparison, we also substitute its generated point clouds with ground-truth point clouds during the inference. 
It is worth emphasizing that TriplaneGaussian also overlooks the challenge of non-uniqueness in Sec.~\ref{sec:pilot}.
In the following sections, we will show that handling this challenge with normal guidance significantly boosts the results.

\subsection{Main Results}
\subsubsection{Novel View Synthesis.}
\label{sec:novel_view}
In this subsection, we conduct novel view synthesis to quantitatively evaluate \namenospace, following the evaluation protocol in Sec.~\ref{sec:data_eval}.
We compare our \name with AGG~\cite{xu2024agg} and TriplaneGaussian~\cite{tri2gsp} in terms of PSNR, SSIM, and LPIPS.
In our experiments, we center-cropped the ground-truth image and the rendered image to a size of $600 \times 600$.
In this way, we significantly reduced the information-less background region so that the evaluation metric could better reflect the rendering quality.
The results listed in Table \ref{table:1} are obtained with the OmniObject3D valid split in Sec.~\ref{sec:data_eval}.
Our method achieves superior performance.
\begin{table}[h!]
\setlength{\tabcolsep}{0.5pt}

\begin{center}
\caption{Comparison with previous methods on the OmniObject3D for Novel View Synthesis.}
\vspace{-10pt}
\label{table:1}
\begin{tabular}{ l|c|c|c  }
 \toprule 
 
 Method & PSNR $\uparrow$ & SSIM $\uparrow$ & LPIPS $\downarrow$ \\
 \hline
 \hline
TriplaneGaussian~\cite{tri2gsp}&   25.8  & 0.881   &0.191\\
 AGG~\cite{xu2024agg}   & 28.4    &0.914&   0.177\\
 \name(ours) & 30.9 & 0.945 &0.134  \\
 \bottomrule
\end{tabular}
\end{center}
\vspace{-10pt}

\end{table}

\begin{figure}[h!]
  \centering
  \includegraphics[width=\linewidth]{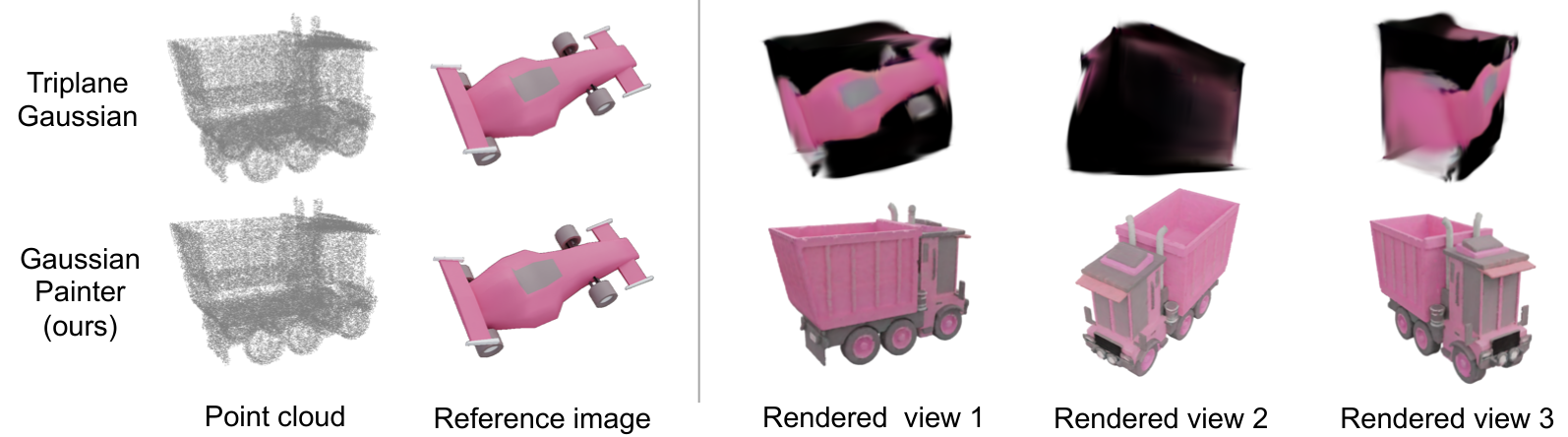}
   \vspace{-10pt}
  \caption{Qualitatively comparison for the task of cross-object appearance transfer on Objaverse. Our method demonstrates the superiority in transferring reasonable appearance to another object. }
  \label{fig:cross_obj}
 \vspace{-5mm}
\end{figure}

\subsubsection{Cross-object Appearance Transfer.}
\label{sec:cross_obj}
Beyond synthesizing novel views, \name shows exciting generalization in transferring the appearance between two quite different objects.
Note that cross-object samples have never been seen by \name during training.
Unlike the novel view synthesis, cross-object transfer requires understanding the semantics of the object parts, in addition to simply learning the mapping from the 2D reference image to the 3D space.
As Fig.~\ref{fig:cross_obj} shows, in the cross-object setting, TriplaneGaussian predicts a large portion of black 3D Gaussians.
We can also find the shape of the formula racing car persists in the rendered truck (first row, rendered view 1).
This phenomenon reveals that TriplaneGaussian may only learn correspondences instead of the correct semantics, resulting in simply ``mapping'' 2D appearance into 3D Gaussians.

\subsection{Ablation Study}
\label{abl}

\subsubsection{Effectiveness of Normal Guidance}
\begin{figure*}[tb]
  \centering
  \includegraphics[width=0.96\linewidth]{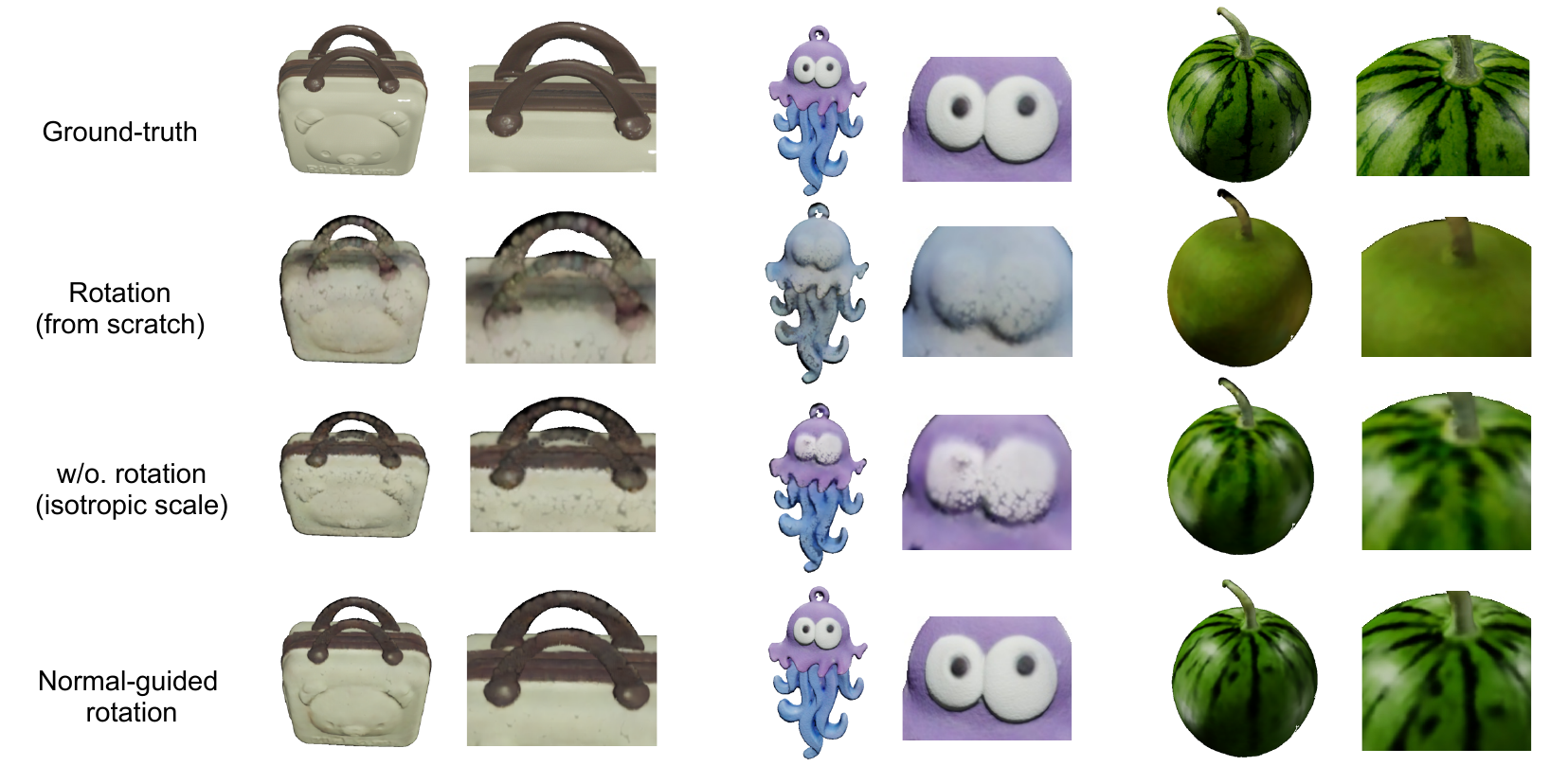}
  \vspace{-2mm}
  \caption{Qualitative comparison between different rotation prediction strategies, using the OmniObject3D validation split constructed in Sec.~\ref{sec:data_eval}.
  To maintain a consistent appearance among different settings for easier comparison, we let the reference view and rendered view be the same.
  In this setting, the model could produce very precise appearance offered by the reference view (i.e., the first row in this figure).}
  \label{fig:demo_normal_guidance}
\end{figure*}

To validate the effectiveness of Normal Guidance, we delete the whole normal prediction branch and add rotation predictions in another branch, which we name as rotation prediction model.
However, training the rotation prediction model from scratch is quite unstable and the model is hard to be well-converged.
Thus, we utilize the pretrained weights of our normal-guided model as initialization and fine-tune the rotation prediction model.
Table~\ref{table:normal_rot} demonstrates the results, which leads to the following findings.
\begin{itemize}
    \item The proposed normal-guided model achieves the best performance, demonstrating the effectiveness of normal guidance.
    \item The rotation prediction model trained from scratch has the worst results, which again verifies the key challenge caused by unstable rotations presented in Sec.~\ref{sec:pilot}.
    \item Normal-guided supervision enables the model to predict stable rotations even without explicitly converting normals to rotations, as demonstrated by the fine-tuned model.
    \item Predicting isotropic Gaussians without rotation leads to stable training but mediocre performance, which is the solution adopted by AGG~\cite{xu2024agg}.
\end{itemize}
We also provide a qualitative comparison between these settings using the objects from OmniObject3D, shown in Fig.~\ref{fig:demo_normal_guidance}.

\begin{table}[t!]

\begin{center}
\vspace{-3mm}
\caption{Comparison between normal guidance and other rotation prediction strategies. Rot. stands for rotation. \dag: this model is fine-tuned to predict rotation from a model trained with normal guidance.}
\label{table:normal_rot}
\resizebox{0.99\linewidth}{!}{
\begin{tabular}{ l|c|c|c  }
 \toprule
 
 Prediction Target&PSNR $\uparrow$ &SSIM $\uparrow$ &LPIPS$\downarrow$\\
 \hline
 \hline
Rot. (from scratch) &  25.8 & 0.899 &0.211 \\
Rot. (fine-tuned)$^\dag$ & 29.5 & 0.924 &0.156  \\
w/o. Rot. (isotropic scale) & 28.4 & 0.914 & 0.177 \\
Normal-guided Rot. (ours) &  30.9 & 0.945 &0.134 \\
 \bottomrule
\end{tabular}}
\end{center}
\vspace{-10pt}

\end{table}

\subsubsection{The Effectiveness of Multiscale Triplane and Cross-modal Attention.}
The multiscale characteristic and cross-model attention (Eq.~(\ref{eq:attn})) are two crucial techniques for building effective triplane representation.
To support our claim, we develop three model variants and compare their performance with the Novel View Synthesis on the OmniObject3D dataset.
The results are listed in Table~\ref{table:attn_tri}.
\begin{table}[t!]
\setlength{\tabcolsep}{1pt}
\begin{center}
\caption{Effectiveness of multi-scale triplane and cross-modal attention (Eq.~(\ref{eq:attn})). The single-scale and multi-scale experiments do not adopt cross-modal attention. The full model integrates multi-scale triplane and cross-modal attention. \dag: The attention-only setting does not obtain information from triplanes. It basically follows Eq.~(\ref{eq:attn}) to interact with images but uses flattened 3D occupancy features as the attention queries.}
\label{table:attn_tri}
\begin{tabular}{ l|c|c|c|c  }
\toprule
 Design Choice&PSNR $\uparrow$ &SSIM $\uparrow$ &LPIPS $\downarrow$& \#Param. \\
 \hline
 \hline
Single-scale triplane&   29.2  & 0.926 &0.155&511 M\\
Multi-scale triplane   & 30.6    &0.930&0.142&549 M\\
Attention-only$^\dag$   & 27.4    &0.905&0.198&458 M\\
Full model & 30.9 & 0.945 &0.134&591 M \\
 \bottomrule
\end{tabular}
\vspace{-2mm}
\end{center}

\end{table}

\begin{table}[!t]
\begin{center}
\caption{Impact of noisy point cloud.}
\vspace{-2mm}
\label{table:noise point cloud}
\resizebox{1.0\columnwidth}{!}{
\begin{tabular}{l|cccccc}
\hline
\hline
Noisy Points & 0\% &10\% & 30\% & 50\%& 70\% & 90\%  \\
\hline
PSNR & 27.5 & 27.1 & 26.9 & 25.9 &24.3&23.7  \\
\hline
\hline
\end{tabular}}
\vspace{-4mm}
\end{center}
\end{table}

\subsubsection{Performence on generated point clouds and noisy point clouds.}
\begin{figure}[!t]
  \centering
  \includegraphics[width=1\columnwidth]{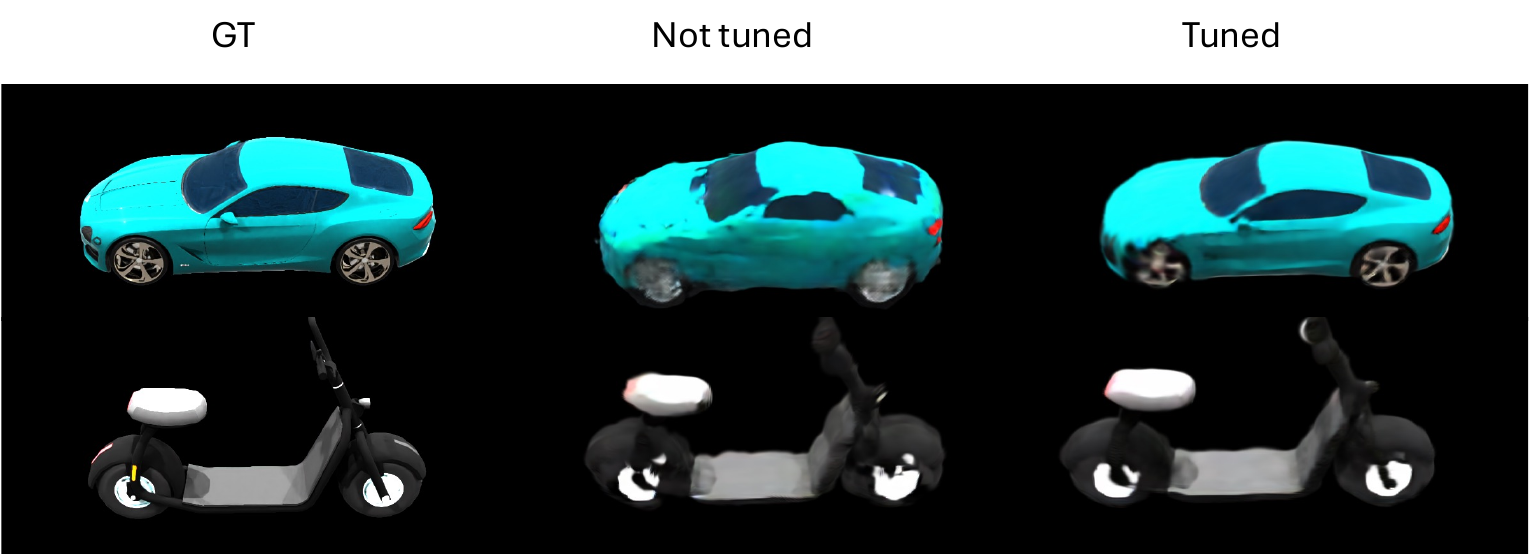}
  \caption{The results of GaussianPainter on generated point cloud for novel view synthesis. The reference image is sampled from a different view.} \label{fig:gen_pcd}
\vspace{-2mm}
\end{figure}

We further conduct experiments to verify how well our method adapts to generate point clouds. We utilize a point cloud generator~\cite{tri2gsp} to generate point clouds, which are then painted into renderable Gaussians with our \name. As shown in Fig. \ref{fig:gen_pcd} (2nd column), our model can directly generalize to generated point clouds. After finetuning our model on the point cloud generator, the results are further improved. More rigorously, we analyze the robustness of our method against noisy point clouds. These point clouds are generated by perturbing a portion of the GT point clouds with Gaussian noise. Table~\ref{table:noise point cloud} shows that our model effectively handles highly noisy input point clouds.

\begin{figure*}[!t]
  \centering
  \includegraphics[width=0.88\linewidth]{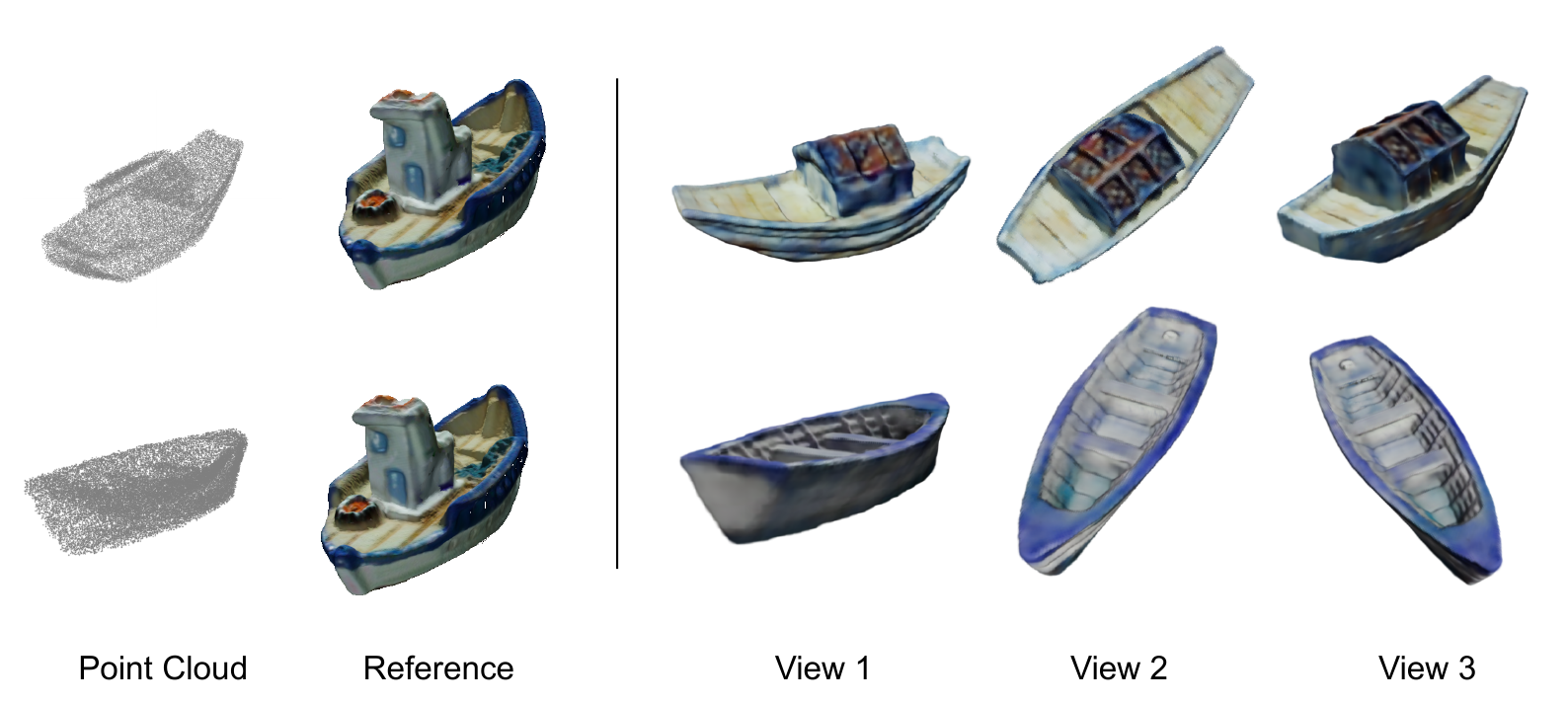}
  \caption{Examples in OmniObject3D. The point clouds and reference images are from different boats.
  Note that the quality of ground-truth objects in OminiObject3D is lower than those in Objaverse, resulting in a relatively lower visual quality of rendered views.
  }
  \label{fig:omni3d}
\end{figure*}

\subsection{Model Analysis} \label{sec:model_analysis}

\textbf{Upper-bound study of rotation representations.}
A potential concern on normal-guided rotation is whether it hinders the capacity of 3D Gaussians since it has a lower degree of freedom than the original rotation parameters in 3DGS.

To address this concern, we train three types of 3D rotation representations for comparison. (1) The unconstrained rotation in original 3D Gaussians. (2) Identity rotation (i.e., no rotation). In this setting, each Gaussian has an isotropic scale, and the rotation matrix is set to identity. (3) Our normal-guided rotation. In this setting, we maintain \emph{two} sets of 3D Gaussians sharing locations. One adopts isotropic Gaussians to optimize the normal, similar to the Isotropic Normal Rendering module in Sec.~\ref{sec:gs_prediction}. 
Another one uses the optimized normals to define rotations following the procedure in Fig.~\ref{fig:n2q} while optimizing the remaining parameters for image rendering. 
As we can see from Table~\ref{table:normal_pp}, our normal-guided rotation only has a slight performance drop compared with the unconstrained rotation, revealing the normal-guided rotation is an acceptable approximation. 
Moreover, the isotropic Gaussian is much worse, which reflects the importance of maintaining an anisotropic design in the proposed \namenospace.

\begin{table}[t!]
\setlength{\tabcolsep}{0.5pt}
\begin{center}
\caption{Upper-bound study to understand the capacity of different rotation representations. The 3D Gaussians are optimized from multi-view images instead of being predicted by our model.}
\label{table:normal_pp}
\begin{tabular}{l|c|c|c}
 \toprule
 Method &PSNR $\uparrow$ &SSIM $\uparrow$ &LPIPS $\downarrow$  \\
 \hline
 \hline
Unconstrained rot. (upper-bound)&   29.16 & 0.9611 &0.133 \\
w/o. rotation (isotropic scale)& 26.30    &0.9373&0.148\\
 Normal-guided rotation (ours) & 28.70 & 0.9566 &0.138 \\
 \bottomrule
\end{tabular}
\end{center}
\vspace{-2mm}
\end{table}

\textbf{Semantic understanding in the texture injection.} 
Although texture injection achieves impressive rendering results, whether it truly understands semantics remains an open question. To further investigate the semantic understanding capability of texture injection, we conducted the following visual experiments on the OmniObject dataset, as shown in Fig.~\ref{fig:omni3d}.

Specifically, we employ a ship as the reference image and perform texture injection on the point clouds of two other ships.
Despite the substantial disparity in geometry between the reference image and the target point clouds, we observe meaningful appearance injection.
In the first row, we observe two notable correspondences between the reference image and the generated ship:

(1) The ship deck region in the rendered image exhibits the same color and texture as the reference image, indicating a potential understanding of the semantics of the deck.

(2) The side and cabinet of the ship in the reference image are mapped with a similar texture and color to the corresponding regions on the generated ship, effectively achieving the desired appearance transfer.

A similar understanding of semantic correspondences can be also found in the second row:

(1) Our model accurately maps the white outside ship body from the reference image to the target point cloud. 

(2) Our model predicts the ship's side to be blue, aligning with the corresponding color in the reference image. 

These results serve as evidence that our model comprehends the semantics to a degree.

\begin{figure*}[th]
  \centering
  \includegraphics[width=0.85\linewidth]{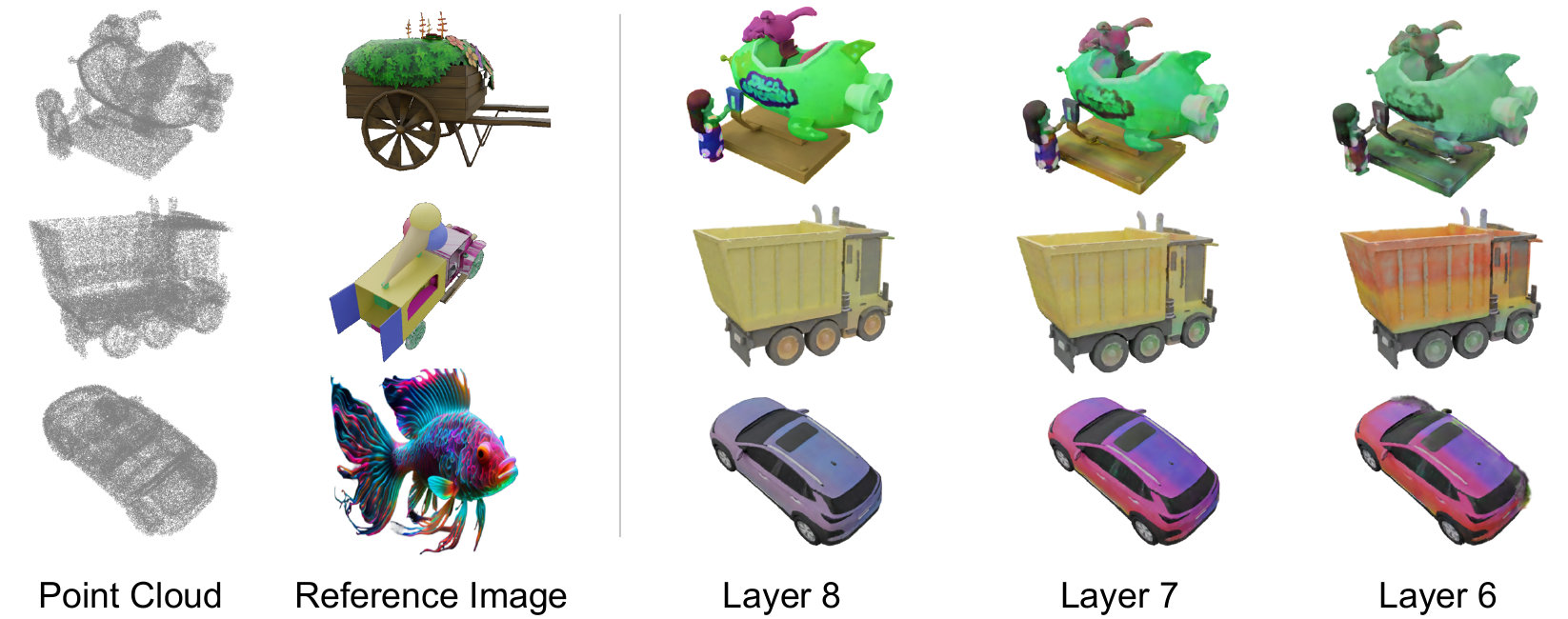}
  \caption{Demonstration using features from different DINOv2 layers. The results show that different visual features could provide diversity for Gaussian painting.}
  \label{fig:dino_layer}
\end{figure*}

\textbf{Diversity from different DINO layers.}
A surprising property of \name is that even trained with the image feature from $8$-th DINO layer, using features from shallower layers during inference still yields reasonable even better results. 
Such property could provide diversity for the painting results while maintaining reasonable overall styles.
For demonstration, we provide examples of cross-object appearance injection using features from different layers in Fig.~\ref{fig:dino_layer}.
The first two cases utilize Objaverse images as references, while the last case involves a web image. 
Although the $6$-th and $7$-th layers of DINOv2 are not directly passed to our triplane module, the texture injection process still achieves successful results.
Notably, the $7$-th layer of DINOv2 exhibits exceptional painting quality while simultaneously showcasing a strong resemblance to the reference appearance.

\begin{figure*}[h!]
  \centering
  \includegraphics[width=0.85\linewidth]{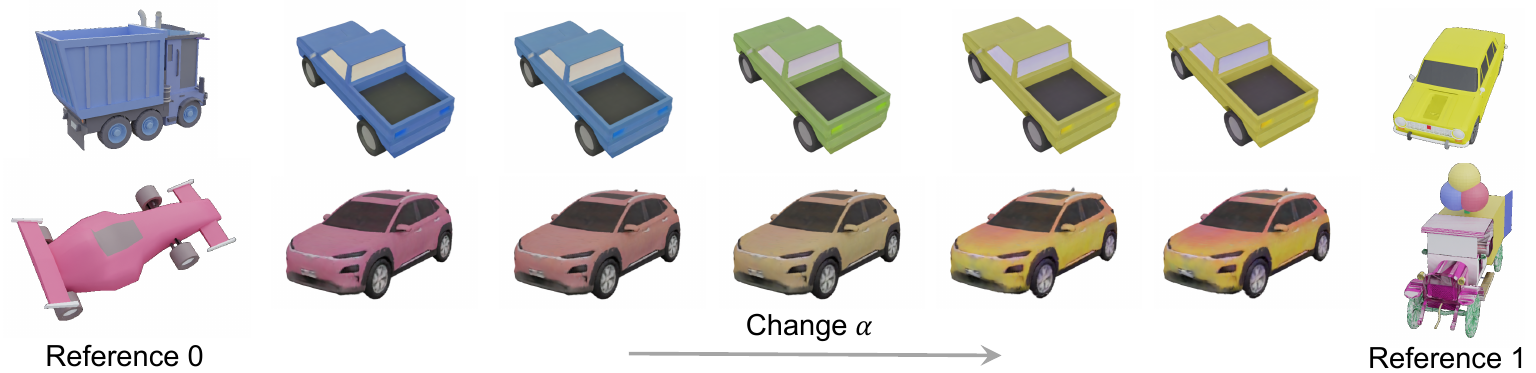}
  \caption{Demonstration of style control with two reference images. An intriguing property is the changing process abides by the color mixing effect in the real world even if the mixup is only applied to features.}
  \label{fig:control}
 \vspace{-2mm}
\end{figure*}

\textbf{Multi-reference Control.} 
Although trained with single reference 
image input, \name is capable of combining the styles of multiple reference images during inference.
We demonstrate this property by feature mixup.
Let $\bm{f}_0,\bm{f}_1$ be two encoded DINOv2 features, we define the mixed visual feature as $\bm{f}_{mix} = \alpha \bm{f}_0+(1-\alpha)\bm{f}_1$. By adjusting $\alpha$, we control the style preference, as Fig.~\ref{fig:control} shows.
This property offers better controllability and diversity of \namenospace.

\section{Conclusion and Future Work}
In this study, we present \namenospace, an innovative method that effectively transforms point clouds into high-quality Gaussian splatting fields in a feed-forward manner, using any reference image with appearance information.
Our approach tackles the non-uniqueness challenge in fitting an anisotropic Gaussian splatting field by introducing normal guidance to stabilize the rotation parameter, significantly reducing the fitting difficulty. Additionally, we introduce a multi-scale triplane representation to better preserve the appearance fidelity of the reference image. \name shows the capacity of high-quality, diverse, and robust 3D content creation from point clouds in a single pass.
\par
However, \name has not been applied to large-scale scene-level point clouds. Our future research endeavors involve scaling the size of our training dataset, scene-level painting, and better controllability by text instructions and intricate appearance.

\bigskip

\bibliography{aaai25}

\begin{thebibliography}{47}
\providecommand{\natexlab}[1]{#1}

\bibitem[{Chen, Wang, and Liu(2023)}]{chen2023text}
Chen, Z.; Wang, F.; and Liu, H. 2023.
\newblock Text-to-3d using gaussian splatting.
\newblock \emph{arXiv preprint arXiv:2309.16585}.

\bibitem[{Chung et~al.(2023)Chung, Lee, Nam, Lee, and Lee}]{chung2023luciddreamer}
Chung, J.; Lee, S.; Nam, H.; Lee, J.; and Lee, K.~M. 2023.
\newblock Luciddreamer: Domain-free generation of 3d gaussian splatting scenes.
\newblock \emph{arXiv preprint arXiv:2311.13384}.

\bibitem[{{Dawson-Haggerty et al.}(2019-12-8)}]{trimesh}
{Dawson-Haggerty et al.} 2019-12-8.
\newblock trimesh.
\newblock In \emph{https://trimesh.org/}, 3.2.0.

\bibitem[{Deitke et~al.(2023)Deitke, Schwenk, Salvador, Weihs, Michel, VanderBilt, Schmidt, Ehsani, Kembhavi, and Farhadi}]{objaverse}
Deitke, M.; Schwenk, D.; Salvador, J.; Weihs, L.; Michel, O.; VanderBilt, E.; Schmidt, L.; Ehsani, K.; Kembhavi, A.; and Farhadi, A. 2023.
\newblock Objaverse: A universe of annotated 3d objects.
\newblock In \emph{CVPR}, 13142--13153.

\bibitem[{Dosovitskiy et~al.(2021)Dosovitskiy, Beyer, Kolesnikov, Weissenborn, Zhai, Unterthiner, Dehghani, Minderer, Heigold, Gelly et~al.}]{vit}
Dosovitskiy, A.; Beyer, L.; Kolesnikov, A.; Weissenborn, D.; Zhai, X.; Unterthiner, T.; Dehghani, M.; Minderer, M.; Heigold, G.; Gelly, S.; et~al. 2021.
\newblock An image is worth 16x16 words: Transformers for image recognition at scale.
\newblock In \emph{ICLR}.

\bibitem[{Gao et~al.(2023)Gao, Gu, Lin, Zhu, Cao, Zhang, and Yao}]{gao2023relightable}
Gao, J.; Gu, C.; Lin, Y.; Zhu, H.; Cao, X.; Zhang, L.; and Yao, Y. 2023.
\newblock Relightable 3d gaussian: Real-time point cloud relighting with brdf decomposition and ray tracing.
\newblock \emph{arXiv preprint arXiv:2311.16043}.

\bibitem[{Gu{\'e}don and Lepetit(2023)}]{guedon2023sugar}
Gu{\'e}don, A.; and Lepetit, V. 2023.
\newblock Sugar: Surface-aligned gaussian splatting for efficient 3d mesh reconstruction and high-quality mesh rendering.
\newblock \emph{arXiv preprint arXiv:2311.12775}.

\bibitem[{Haque et~al.(2023)Haque, Tancik, Efros, Holynski, and Kanazawa}]{haque2023instruct}
Haque, A.; Tancik, M.; Efros, A.~A.; Holynski, A.; and Kanazawa, A. 2023.
\newblock Instruct-nerf2nerf: Editing 3d scenes with instructions.
\newblock \emph{ICCV}.

\bibitem[{He et~al.(2024)He, Ma, Huang, Huang, Gao, Wei, Dai, Han, and Liu}]{he2024freeedit}
He, R.; Ma, K.; Huang, L.; Huang, S.; Gao, J.; Wei, X.; Dai, J.; Han, J.; and Liu, S. 2024.
\newblock Freeedit: Mask-free reference-based image editing with multi-modal instruction.
\newblock \emph{arXiv preprint arXiv:2409.18071}.

\bibitem[{Huang et~al.(2025)Huang, Fang, Zhang, Song, Liu, Liu, and Li}]{huang2025fouriscale}
Huang, L.; Fang, R.; Zhang, A.; Song, G.; Liu, S.; Liu, Y.; and Li, H. 2025.
\newblock Fouriscale: A frequency perspective on training-free high-resolution image synthesis.
\newblock In \emph{European Conference on Computer Vision}, 196--212. Springer.

\bibitem[{Kerbl et~al.(2023)Kerbl, Kopanas, Leimk{\"u}hler, and Drettakis}]{gsp}
Kerbl, B.; Kopanas, G.; Leimk{\"u}hler, T.; and Drettakis, G. 2023.
\newblock 3D Gaussian Splatting for Real-Time Radiance Field Rendering.
\newblock \emph{ACM Transactions on Graphics}, 42(4).

\bibitem[{Kondo et~al.(2021)Kondo, Ikeda, Tagliasacchi, Matsuo, Ochiai, and Gu}]{kondo2021vaxnerf}
Kondo, N.; Ikeda, Y.; Tagliasacchi, A.; Matsuo, Y.; Ochiai, Y.; and Gu, S.~S. 2021.
\newblock Vaxnerf: Revisiting the classic for voxel-accelerated neural radiance field.
\newblock \emph{arXiv preprint arXiv:2111.13112}.

\bibitem[{Lassner and Zollhofer(2021)}]{lassner2021pulsar}
Lassner, C.; and Zollhofer, M. 2021.
\newblock Pulsar: Efficient sphere-based neural rendering.
\newblock In \emph{CVPR}, 1440--1449.

\bibitem[{Li et~al.(2021)Li, Feng, She, Ding, Wang, and Lee}]{li2021mine}
Li, J.; Feng, Z.; She, Q.; Ding, H.; Wang, C.; and Lee, G.~H. 2021.
\newblock Mine: Towards continuous depth mpi with nerf for novel view synthesis.
\newblock In \emph{ICCV}, 12578--12588.

\bibitem[{Li, Wang, and Tseng(2023)}]{li2023gaussiandiffusion}
Li, X.; Wang, H.; and Tseng, K.-K. 2023.
\newblock Gaussiandiffusion: 3d gaussian splatting for denoising diffusion probabilistic models with structured noise.
\newblock \emph{arXiv preprint arXiv:2311.11221}.

\bibitem[{Liang et~al.(2023)Liang, Yang, Lin, Li, Xu, and Chen}]{liang2023luciddreamer}
Liang, Y.; Yang, X.; Lin, J.; Li, H.; Xu, X.; and Chen, Y. 2023.
\newblock Luciddreamer: Towards high-fidelity text-to-3d generation via interval score matching.
\newblock \emph{arXiv preprint arXiv:2311.11284}.

\bibitem[{Lin et~al.(2023)Lin, Gao, Tang, Takikawa, Zeng, Huang, Kreis, Fidler, Liu, and Lin}]{lin2023magic3d}
Lin, C.-H.; Gao, J.; Tang, L.; Takikawa, T.; Zeng, X.; Huang, X.; Kreis, K.; Fidler, S.; Liu, M.-Y.; and Lin, T.-Y. 2023.
\newblock Magic3d: High-resolution text-to-3d content creation.
\newblock In \emph{CVPR}, 300--309.

\bibitem[{Lionar et~al.(2024)Lionar, Xu, Lin, and Lee}]{sdf}
Lionar, S.; Xu, X.; Lin, M.; and Lee, G.~H. 2024.
\newblock Nu-mcc: Multiview compressive coding with neighborhood decoder and repulsive udf.
\newblock \emph{NeurIPS}, 36.

\bibitem[{Liu et~al.(2024)Liu, Xu, Jin, Chen, Varma~T, Xu, and Su}]{liu2024one}
Liu, M.; Xu, C.; Jin, H.; Chen, L.; Varma~T, M.; Xu, Z.; and Su, H. 2024.
\newblock One-2-3-45: Any single image to 3d mesh in 45 seconds without per-shape optimization.
\newblock \emph{NeurIPS}, 36.

\bibitem[{Liu et~al.(2023{\natexlab{a}})Liu, Wu, Hoorick, Tokmakov, Zakharov, and Vondrick}]{liu2023zero1to3}
Liu, R.; Wu, R.; Hoorick, B.~V.; Tokmakov, P.; Zakharov, S.; and Vondrick, C. 2023{\natexlab{a}}.
\newblock Zero-1-to-3: Zero-shot One Image to 3D Object.
\newblock In \emph{ICCV}.

\bibitem[{Liu et~al.(2023{\natexlab{b}})Liu, Zhan, Tang, Shan, Zeng, Lin, Liu, and Liu}]{liu2023humangaussian}
Liu, X.; Zhan, X.; Tang, J.; Shan, Y.; Zeng, G.; Lin, D.; Liu, X.; and Liu, Z. 2023{\natexlab{b}}.
\newblock Humangaussian: Text-driven 3d human generation with gaussian splatting.
\newblock \emph{arXiv preprint arXiv:2311.17061}.

\bibitem[{Luiten et~al.(2023)Luiten, Kopanas, Leibe, and Ramanan}]{luiten2023dynamic}
Luiten, J.; Kopanas, G.; Leibe, B.; and Ramanan, D. 2023.
\newblock Dynamic 3d gaussians: Tracking by persistent dynamic view synthesis.
\newblock \emph{arXiv preprint arXiv:2308.09713}.

\bibitem[{Melas-Kyriazi et~al.(2023)Melas-Kyriazi, Laina, Rupprecht, and Vedaldi}]{melas2023realfusion}
Melas-Kyriazi, L.; Laina, I.; Rupprecht, C.; and Vedaldi, A. 2023.
\newblock Realfusion: 360deg reconstruction of any object from a single image.
\newblock In \emph{CVPR}, 8446--8455.

\bibitem[{Mildenhall et~al.(2021)Mildenhall, Srinivasan, Tancik, Barron, Ramamoorthi, and Ng}]{nerf}
Mildenhall, B.; Srinivasan, P.~P.; Tancik, M.; Barron, J.~T.; Ramamoorthi, R.; and Ng, R. 2021.
\newblock Nerf: Representing scenes as neural radiance fields for view synthesis.
\newblock \emph{Communications of the ACM}, 65(1): 99--106.

\bibitem[{Nichol et~al.(2022)Nichol, Jun, Dhariwal, Mishkin, and Chen}]{nichol2022point}
Nichol, A.; Jun, H.; Dhariwal, P.; Mishkin, P.; and Chen, M. 2022.
\newblock Point-e: A system for generating 3d point clouds from complex prompts.
\newblock \emph{arXiv preprint arXiv:2212.08751}.

\bibitem[{Oquab et~al.(2023)Oquab, Darcet, Moutakanni, Vo, Szafraniec, Khalidov, Fernandez, Haziza, Massa, El-Nouby et~al.}]{oquab2023dinov2}
Oquab, M.; Darcet, T.; Moutakanni, T.; Vo, H.; Szafraniec, M.; Khalidov, V.; Fernandez, P.; Haziza, D.; Massa, F.; El-Nouby, A.; et~al. 2023.
\newblock Dinov2: Learning robust visual features without supervision.
\newblock \emph{arXiv preprint arXiv:2304.07193}.

\bibitem[{Park et~al.(2019)Park, Florence, Straub, Newcombe, and Lovegrove}]{park2019deepsdf}
Park, J.~J.; Florence, P.; Straub, J.; Newcombe, R.; and Lovegrove, S. 2019.
\newblock Deepsdf: Learning continuous signed distance functions for shape representation.
\newblock In \emph{CVPR}, 165--174.

\bibitem[{Poole et~al.(2022)Poole, Jain, Barron, and Mildenhall}]{poole2022dreamfusion}
Poole, B.; Jain, A.; Barron, J.~T.; and Mildenhall, B. 2022.
\newblock Dreamfusion: Text-to-3d using 2d diffusion.
\newblock \emph{arXiv preprint arXiv:2209.14988}.

\bibitem[{Qian et~al.(2023)Qian, Wang, Mihajlovic, Geiger, and Tang}]{qian20233dgs}
Qian, Z.; Wang, S.; Mihajlovic, M.; Geiger, A.; and Tang, S. 2023.
\newblock 3dgs-avatar: Animatable avatars via deformable 3d gaussian splatting.
\newblock \emph{arXiv preprint arXiv:2312.09228}.

\bibitem[{Rombach et~al.(2022)Rombach, Blattmann, Lorenz, Esser, and Ommer}]{ldm}
Rombach, R.; Blattmann, A.; Lorenz, D.; Esser, P.; and Ommer, B. 2022.
\newblock High-resolution image synthesis with latent diffusion models.
\newblock In \emph{CVPR}, 10684--10695.

\bibitem[{Ronneberger, Fischer, and Brox(2015)}]{unet}
Ronneberger, O.; Fischer, P.; and Brox, T. 2015.
\newblock U-net: Convolutional networks for biomedical image segmentation.
\newblock In \emph{MICCAI}, 234--241. Springer.

\bibitem[{Shih et~al.(2020)Shih, Su, Kopf, and Huang}]{shih20203d}
Shih, M.-L.; Su, S.-Y.; Kopf, J.; and Huang, J.-B. 2020.
\newblock 3d photography using context-aware layered depth inpainting.
\newblock In \emph{CVPR}, 8028--8038.

\bibitem[{Sun et~al.(2020)Sun, Wang, Liu, Siegel, and Sarma}]{sun2020pointgrow}
Sun, Y.; Wang, Y.; Liu, Z.; Siegel, J.; and Sarma, S. 2020.
\newblock Pointgrow: Autoregressively learned point cloud generation with self-attention.
\newblock In \emph{WACV}, 61--70.

\bibitem[{Tang et~al.(2024)Tang, Ren, Zhou, Liu, and Zeng}]{tang2023dreamgaussian}
Tang, J.; Ren, J.; Zhou, H.; Liu, Z.; and Zeng, G. 2024.
\newblock Dreamgaussian: Generative gaussian splatting for efficient 3d content creation.
\newblock \emph{ICLR}.

\bibitem[{Wang et~al.(2023)Wang, Du, Li, Yeh, and Shakhnarovich}]{wang2023score}
Wang, H.; Du, X.; Li, J.; Yeh, R.~A.; and Shakhnarovich, G. 2023.
\newblock Score jacobian chaining: Lifting pretrained 2d diffusion models for 3d generation.
\newblock In \emph{CVPR}, 12619--12629.

\bibitem[{Wang et~al.(2024)Wang, Lu, Wang, Bao, Li, Su, and Zhu}]{wang2024prolificdreamer}
Wang, Z.; Lu, C.; Wang, Y.; Bao, F.; Li, C.; Su, H.; and Zhu, J. 2024.
\newblock Prolificdreamer: High-fidelity and diverse text-to-3d generation with variational score distillation.
\newblock \emph{NeurIPS}, 36.

\bibitem[{Wu et~al.(2023)Wu, Zhang, Fu, Wang, Ren, Pan, Wu, Yang, Wang, Qian et~al.}]{ommni3d}
Wu, T.; Zhang, J.; Fu, X.; Wang, Y.; Ren, J.; Pan, L.; Wu, W.; Yang, L.; Wang, J.; Qian, C.; et~al. 2023.
\newblock Omniobject3d: Large-vocabulary 3d object dataset for realistic perception, reconstruction and generation.
\newblock In \emph{CVPR}, 803--814.

\bibitem[{Xu et~al.(2022{\natexlab{a}})Xu, Jiang, Wang, Fan, Shi, and Wang}]{xu2022sinnerf}
Xu, D.; Jiang, Y.; Wang, P.; Fan, Z.; Shi, H.; and Wang, Z. 2022{\natexlab{a}}.
\newblock Sinnerf: Training neural radiance fields on complex scenes from a single image.
\newblock In \emph{ECCV}, 736--753. Springer.

\bibitem[{Xu et~al.(2024)Xu, Yuan, Mardani, Liu, Song, Wang, and Vahdat}]{xu2024agg}
Xu, D.; Yuan, Y.; Mardani, M.; Liu, S.; Song, J.; Wang, Z.; and Vahdat, A. 2024.
\newblock Agg: Amortized generative 3d gaussians for single image to 3d.
\newblock \emph{arXiv preprint arXiv:2401.04099}.

\bibitem[{Xu et~al.(2022{\natexlab{b}})Xu, Xu, Philip, Bi, Shu, Sunkavalli, and Neumann}]{xu2022point}
Xu, Q.; Xu, Z.; Philip, J.; Bi, S.; Shu, Z.; Sunkavalli, K.; and Neumann, U. 2022{\natexlab{b}}.
\newblock Point-nerf: Point-based neural radiance fields.
\newblock In \emph{CVPR}, 5438--5448.

\bibitem[{Yan et~al.(2024)Yan, Lin, Zhou, Wang, Sun, Zhan, Lang, Zhou, and Peng}]{yan2024street}
Yan, Y.; Lin, H.; Zhou, C.; Wang, W.; Sun, H.; Zhan, K.; Lang, X.; Zhou, X.; and Peng, S. 2024.
\newblock Street gaussians for modeling dynamic urban scenes.
\newblock \emph{arXiv preprint arXiv:2401.01339}.

\bibitem[{Yang et~al.(2019)Yang, Huang, Hao, Liu, Belongie, and Hariharan}]{yang2019pointflow}
Yang, G.; Huang, X.; Hao, Z.; Liu, M.-Y.; Belongie, S.; and Hariharan, B. 2019.
\newblock Pointflow: 3d point cloud generation with continuous normalizing flows.
\newblock In \emph{ICCV}, 4541--4550.

\bibitem[{Yifan et~al.(2019)Yifan, Serena, Wu, {\"O}ztireli, and Sorkine-Hornung}]{yifan2019differentiable}
Yifan, W.; Serena, F.; Wu, S.; {\"O}ztireli, C.; and Sorkine-Hornung, O. 2019.
\newblock Differentiable surface splatting for point-based geometry processing.
\newblock \emph{ACM Transactions on Graphics (TOG)}, 38(6): 1--14.

\bibitem[{Yugay et~al.(2023)Yugay, Li, Gevers, and Oswald}]{yugay2023gaussian}
Yugay, V.; Li, Y.; Gevers, T.; and Oswald, M.~R. 2023.
\newblock Gaussian-slam: Photo-realistic dense slam with gaussian splatting.
\newblock \emph{arXiv preprint arXiv:2312.10070}.

\bibitem[{Zhang et~al.(2023)Zhang, Wei, Jiang, Zhang, Zuo, and Tian}]{zhang2023controlvideo}
Zhang, Y.; Wei, Y.; Jiang, D.; Zhang, X.; Zuo, W.; and Tian, Q. 2023.
\newblock Controlvideo: Training-free controllable text-to-video generation.
\newblock \emph{arXiv preprint arXiv:2305.13077}.

\bibitem[{Zhou et~al.(2023)Zhou, Lin, Shan, Wang, Sun, and Yang}]{zhou2023drivinggaussian}
Zhou, X.; Lin, Z.; Shan, X.; Wang, Y.; Sun, D.; and Yang, M.-H. 2023.
\newblock Drivinggaussian: Composite gaussian splatting for surrounding dynamic autonomous driving scenes.
\newblock \emph{arXiv preprint arXiv:2312.07920}.

\bibitem[{Zou et~al.(2023)Zou, Yu, Guo, Li, Liang, Cao, and Zhang}]{tri2gsp}
Zou, Z.-X.; Yu, Z.; Guo, Y.-C.; Li, Y.; Liang, D.; Cao, Y.-P.; and Zhang, S.-H. 2023.
\newblock Triplane meets gaussian splatting: Fast and generalizable single-view 3d reconstruction with transformers.
\newblock \emph{arXiv preprint arXiv:2312.09147}.

\end{thebibliography}

\end{document}